\documentclass[11pt]{article}

\usepackage{graphicx}
\usepackage{subfigure} 
\usepackage[round]{natbib}
\usepackage{amsmath}
\usepackage{mathtools}
\usepackage[top=2.5cm,bottom=2.5cm,left=2.5cm,right=2.5cm]{geometry}

\newcommand{\vect}[1]{\mathbf{x}}
\newcommand{\vx}[0]{\vect{x}}
\newcommand{\erf}{\operatorname{erf}}
\newcommand{\erfinv}{\operatorname{erfinv}}

\newcommand{\ellbound}{a_\ell}

\title{Gaussian process modelling of multiple short time series}

\author{
Hande Topa$^1$
and
Antti Honkela$^2$ \\[1em]
\small $^1$ Helsinki Institute for Information Technology HIIT\\
\small Department of Information and Computer Science\\
\small Aalto University, Helsinki, Finland\\
\small \texttt{hande.topa@aalto.fi}\\
\small $^2$ Helsinki Institute for Information Technology HIIT\\
\small Department of Computer Science\\
\small University of Helsinki, Helsinki, Finland\\
\small \texttt{antti.honkela@hiit.fi}%
}
\date{}

\begin{document} 

\maketitle

\begin{abstract} 
  We present techniques for effective Gaussian process (GP) modelling
  of multiple short time series. These problems are common when applying
  GP models independently to each gene in a gene expression time
  series data set. Such sets typically contain very few time points.
  Naive application of common GP modelling techniques can lead to
  severe over-fitting or under-fitting in a significant fraction of
  the fitted models, depending on the details of the data set.  We
  propose avoiding over-fitting by constraining the GP length-scale to
  values that focus most of the energy spectrum to frequencies
  below the Nyquist frequency corresponding to the sampling frequency
  in the data set.  Under-fitting can be avoided by more informative
  priors on observation noise.  Combining these methods allows
  applying GP methods reliably automatically to large numbers of
  independent instances of short time series.  This is illustrated
  with experiments with both synthetic data and real gene expression
  data.
\end{abstract} 

\section{Introduction}
\label{Introduction}

Gaussian processes (GPs) are a widely applied non-parametric
probabilistic model for continuous data~\citep{Rasmussen:book06}.
Because of their non-parametric nature, they can flexibly adapt to
differently sized data sets and can easily accommodate for example
non-uniformly sampled data.  GPs are computationally very convenient,
because they permit exact marginalisation of the latent process in
regression with a Gaussian likelihood.

Most methods development work on GPs in machine learning has focused
on developing efficient inference for large data
sets~\citep[see, e.g.][]{Quinonero-Candela2005,Snelson:pseudo05,Rasmussen:book06,Titsias:variational09}.
This is an important area, as naive inference algorithms suffer from
cubic computational complexity with respect to the data set size, and
the recently developed methods can successfully reduce this
significantly.

In this paper we focus instead on the other frontier of GP
applications in data sets with a very large number of small independent
instances.  GPs for such applications have recently gathered significant
interest in computational systems biology, where they provide a very
powerful model for sparsely and often irregularly sampled gene
expression time
series~\citep{Lawrence2007,Gao2008,Kirk2009,Stegle2010a,Liu2010,Honkela2010,Kalaitzis2011,Cooke2011,Titsias2012}.
Reliable fitting of very large number of independent models is
important in many applications of these models, such as ranking of
targets of gene regulators~\citep{Honkela2010}.

Most gene expression time series are very short with a great majority
having less than 9 time points~\citep{Ernst2005}, so computational
complexity of any GP inference method will typically not be an
issue. Instead, the application of GP methods in these problems will
face other problems due to lack of and sparseness of data. Depending
on the specific problem, this can easily lead to either over-fitting or
under-fitting. When fitting the models automatically to a large number,
possibly several thousands, of instances, it is impractical to
manually locate and fix these problematic fits.  In this paper we
present methods for setting constraints or more restrictive priors to
some model parameters that help avoid these phenomena.  Earlier
heuristic variants of the length-scale bound have been applied in some
previous works~\citep{Honkela2011,Titsias2012} without detailed
justification, but here we present a new rigorous derivation for the
bound.

\begin{figure*}[tb]
\centering
  \subfigcapskip-0.5ex
  \subfigcapmargin1ex
  \subfigure[The data]{
    \includegraphics[width=0.36\textwidth,trim=23mm 71mm 23mm 71mm,clip]{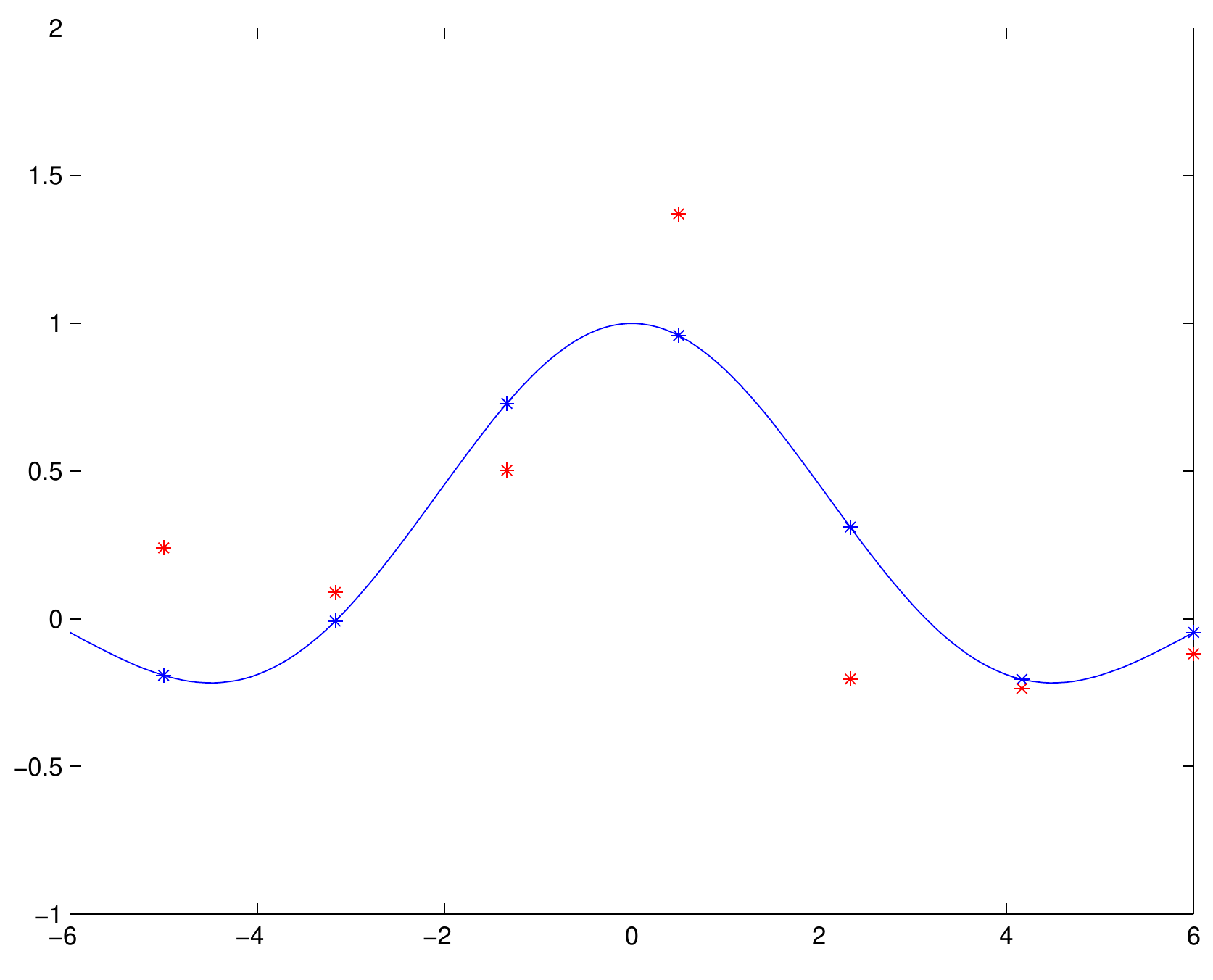}
  }%
  \subfigure[The log-likelihood surface]{
    \includegraphics[width=0.36\textwidth]{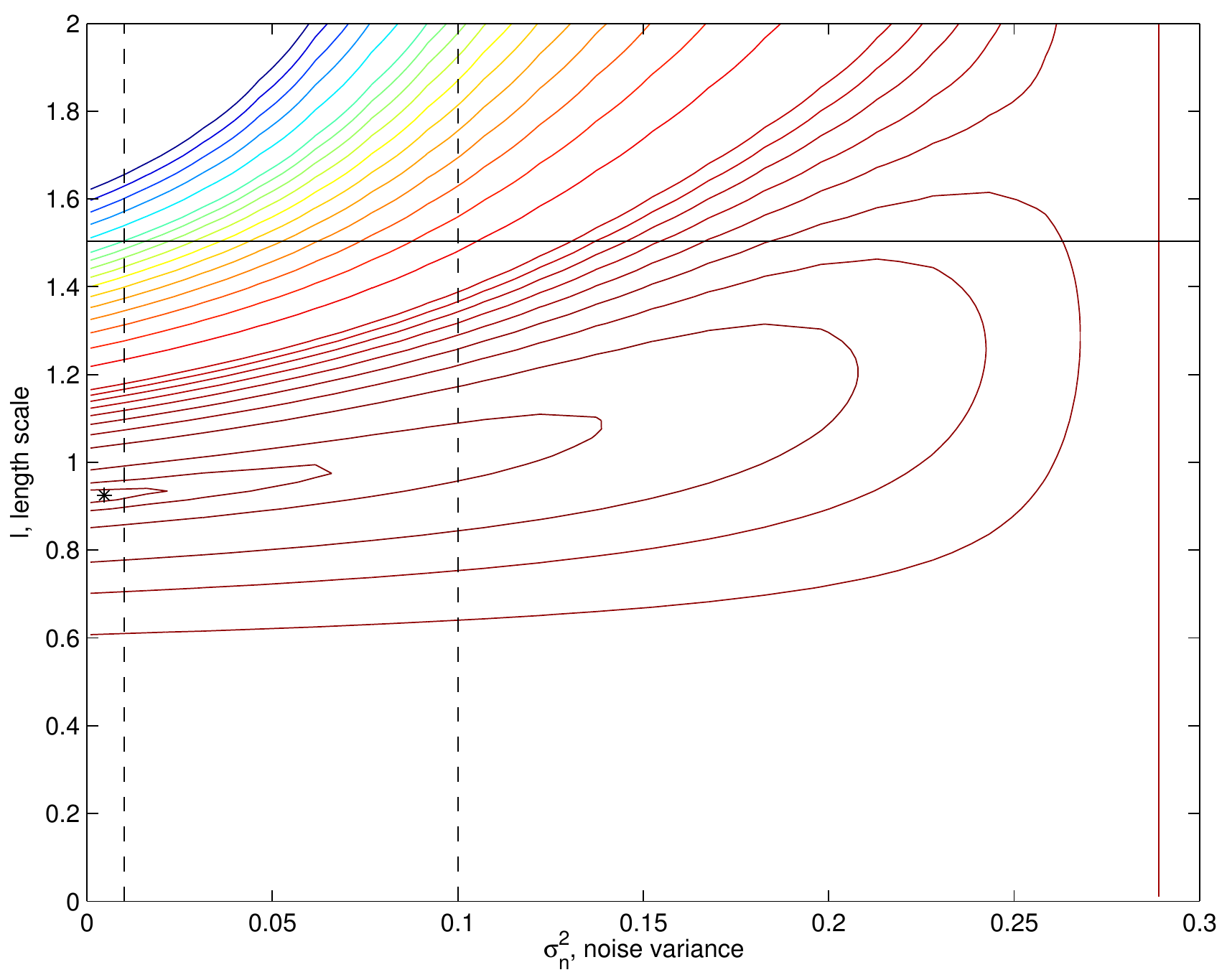}
  }
\caption{
  (a) The underlying function (blue line) for the synthetic example and an
  example data set instance (red dots).
  (b) Contour plot of the log-likelihood surface for different
  $\ell$ and $\sigma^2_n$ values corresponding to the data set
  in (a) showing the maximum likelihood
  solution (black dot) with a very small $\sigma_n^2 \approx 0.005$ and $\ell \approx 0.9$, when no bounds are introduced.}
\label{fig:likelihood_surface}
\end{figure*} 

\section{GP modelling methods for small data sets}
\label{GP modelling methods for small data sets}

A GP is a stochastic process $\{ f(\vx) | \vx \in \mathcal{X}\}$ for
which the marginal distribution at any finite sub-collection of points
$\vx_1, \dots, \vx_n$ is multivariate
Gaussian~\citep{Rasmussen:book06}.  The process is completely defined
by the mean function $\mu(\vx)$ and the covariance function $k(\vx,
\vx')$, that also define the mean vector and covariance matrix of the
multivariate Gaussian over the sub-collection.  For simplicity, we
assume the mean function is identically zero $\mu(\vx) \equiv 0$.

The most widely used covariance function for GPs in machine learning
is the squared exponential covariance~\citep{Rasmussen:book06}
\begin{equation}
  \label{eq:rbf_covariance}
  k_{\text{SE}}(\vx, \vx') = \sigma_f^2
  \exp\left(-\frac{r^2}{2 \ell^2}\right),
\end{equation}
where $r=||\vx - \vx'||$.  The covariance depends on two positive
hyperparameters: variance $\sigma_f^2$ and length-scale $\ell$.  The
squared exponential covariance is infinitely smooth, which is often
too strong an assumption.  A simple generalisation is given by the
Mat\'{e}rn class covariance functions
\begin{equation}
  \label{eq:matern_covariance}
  k_{\text{Matern}}(\vx, \vx') = \sigma_f^2
  \frac{2^{1-\nu}}{\Gamma(\nu)} \left(\frac{\sqrt{2\nu}r}{\ell}\right)^\nu 
  K_\nu \left(\frac{\sqrt{2\nu} r}{\ell}\right),
\end{equation}
with additional positive hyperparameter $\nu$, where $K_\nu$ is a
modified Bessel function~\citep{Rasmussen:book06}.

Both these covariance functions have a length-scale parameter $\ell$
that governs the range of dependencies in the process. A short length
scale corresponds to rapidly varying functions with weak long-range
dependencies, while a large length-scale corresponds to slowly varying
functions. Extremely small length-scale may lead to a situation where
each observation is treated as essentially independent, which makes
the model over-fit.

\subsection{Illustrative Example}
\label{sec:illustrative-example}

We illustrate fitting GPs to small data sets with synthetic data
generated from a
$\mathrm{sinc}(x)$ function, that is, $f(x)=\frac{\sin(x)}{x}$, by
uniformly sampling 7 data points in the interval [-5,6] with a noise
term which is normally distributed with mean $0$ and variance $0.09$.  An example of such a data set is shown in
Fig.~\ref{fig:likelihood_surface}(a).

\begin{figure}[tb]
\centering
\includegraphics[width=0.5\textwidth,trim=10mm 71mm 10mm 71mm,clip]{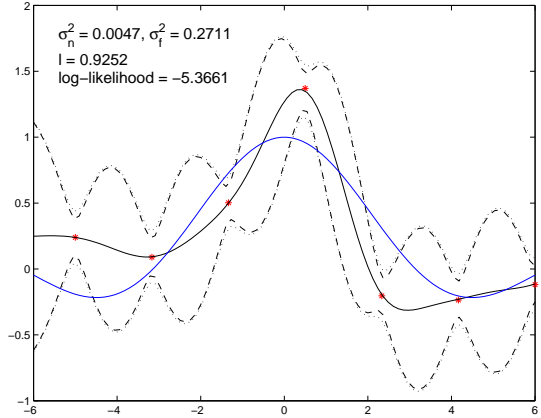}
\caption{A GP model fitted to 7 data points generated from
  $f(x)=\frac{\sin(x)}{x}$ showing over-fitting to the data.  The plot
  shows the posterior mean of the GP (black solid line) together with two
  standard deviation posterior credible regions both for the squared
  exponential covariance (dashed line) and squared exponential plus
  noise covariance (dotted line).  The two posterior credible regions
  are very close to each other, indicating a very small estimated
  observation noise level.}
\label{fig:overfitting}
\end{figure}

Fitting a GP model with squared exponential covariance to the data set
shown in Fig.~\ref{fig:likelihood_surface}(a) and selecting the
maximum likelihood (ML) solution for the squared exponential
covariance variance $\sigma_f^2$ leads to a likelihood surface for length scale
$\ell$ and observation noise variance $\sigma_n^2$ shown in
Fig.~\ref{fig:likelihood_surface}(b).
The ML estimate for the noise is clearly much smaller than the
generative value indicating severe over-fitting to the data.  This
corresponds to a fairly small value for the length-scale compared to
the sampling rate in the data.  The GP model corresponding to the ML
fit is shown in Fig.~\ref{fig:overfitting}.

\subsection{Length-scale bounds}

As seen above,
naive application of common GP modelling techniques can lead to severe
over-fitting or under-fitting, depending on the details of the data
set.
We propose avoiding over-fitting by constraining the GP length-scale $\ell$ to
values that focus most of the energy spectrum to frequencies below the
Nyquist frequency corresponding to the sampling in the data set.
According to the Nyquist sampling
theorem, the Nyquist frequency $f_n=\frac{1}{2\Delta t}$
is the maximal frequency that can be identified in the spectral
representation of the sampled signal~\citep{Tick1966, Nyquist}.  Here
$\Delta t$ is the sampling interval in the data set.
In case of non-uniformly sampled data, we define $\Delta t$
conservatively as the shortest distance between consecutive data
points to obtain the least restrictive bound.

In case of the squared exponential covariance function, the
spectral density is
given by
\begin{equation}
  S_{\text{SE}}(s)= (2\pi \ell^2)^{D/2} \exp(-2\pi^2 \ell^2 s^2),
  \label{eq:se_spec_dens}
\end{equation}
where $D$ is the number of dimensions and $s$ denotes the frequency
\citep{Rasmussen:book06}.
For $D=1$, the corresponding lower bound for the length-scale
can be found by solving the
inequality
\begin{equation}
  \label{eq:se_integ}
  \int\limits_{-\frac{1}{2\Delta t}}^{\frac{1}{2\Delta t}} S_{\text{SE}}(s)\;\mathrm{d}s
  = \erf\left( \frac{\pi \ell}{\sqrt{2} \Delta t} \right)
  \geq \alpha ,
\end{equation}
where $\alpha$ denotes the fraction of the system's energy on the frequencies that are below the Nyquist frequency.

Solving for $\ell$ and setting $\alpha$ to $0.99$, we can obtain the lower bound for the length scale parameter that would constrain at least $99\%$ of the process's energy on the frequencies which are below the Nyquist frequency: 
\begin{equation}
  \ell \geq \ellbound(\alpha) = \frac{\sqrt{2} \erfinv(\alpha)}{\pi} \Delta t
  \approx 0.8199 \times \Delta t,
  \label{l_bound}
\end{equation}
where $\erfinv(x)$ denotes the inverse of the error function. 

For the 1-dimensional Mat\'ern covariance function the corresponding
spectral density is
\begin{equation}
  \label{eq:matern_spec_dens}
  S_{\text{Matern}}(s) = \frac{2\sqrt{\pi} \gamma(\nu+\frac{1}{2}) (2\nu)^\nu}{\gamma(\nu) \ell^{2\nu}}
  \left( \frac{2\nu}{\ell^2} + 4 \pi^2 s^2 \right)^{-(\nu+\frac{1}{2})},
\end{equation}
from which we can derive the fraction of energy in the frequency
interval $\left[-1/2\Delta t, 1/2\Delta t\right]$ as
\begin{equation}
  \label{eq:matern_energy}
  \int\limits_{-\frac{1}{2\Delta t}}^{\frac{1}{2\Delta t}} S_{\text{Matern}}(s)\;\mathrm{d}s 
  = \frac{4\ell \sqrt{2\pi} \gamma(\nu+1/2)}{\Delta t \sqrt{v} \gamma(\nu)}
  \prescript{}{2}F_1\left(\frac{1}{2}, \nu + \frac{1}{2}, \frac{3}{2},
    -\frac{\ell^2 \pi^2}{2 \nu (\Delta t)^2} \right),
%                                                                       2   2  2
%                             1                        1  1      3  -2 l  Pi  t
%      2 l Sqrt[2 Pi] t Gamma[- + v] Hypergeometric2F1[-, - + v, -, ------------]
%                             2                        2  2      2       v
% >    --------------------------------------------------------------------------
%                                   Sqrt[v] Gamma[v]
\end{equation}
where $\prescript{}{2}F_1(a, b, c, x)$ is the hypergeometric
function~\citep{Abramowitz:handbook65}.  Given a fixed value of $\nu$,
appropriate lower bound $\ellbound{}$ for $\ell$ can be derived from
Eq.~(\ref{eq:matern_energy}) using numerical optimisation.

For Bayesian parameter estimation, a uniform prior over the length
scale over the interval $[\ellbound{}, t_n - t_1]$ seems like a reasonable
objective prior.

\subsection{Variance prior}

Bounding the length-scale as above can avoid most obvious
over-fitting, but naive estimation of observation noise will still
frequently lead to mis-estimation of the noise and hence effectively
over-fitting or under-fitting the data.  The easiest way to avoid this is to use
informative priors on the noise that focus the distribution away from
implausibly small and large values.

For gene expression data, one can for example use posterior variances
of the inferred expression levels from pre-processing both for
microarrays~\citep{Liu2005,Du2008} and
RNA-sequencing~\citep{Turro2011,Glaus2012}.  This kind of approach has
been applied e.g.\
in~\citep{Lawrence2007,Gao2008,Honkela2010,Titsias2012}.

As an alternative, \citet{Cooke2011} use an
empirical variance estimate from multiple replicates as the approximate
lower bound
and total variance as the approximate upper bound for a semi-empirical prior on
the variance.

\section{Experiments}
\label{Experiments}

We present experimental results highlighting the over-fitting caused
by bad length-scale estimates on synthetic data and real gene
expression data.

\begin{figure*}[tb]
  \centering
  \subfigcapskip-0.5ex
  \subfigcapmargin1ex
  \subfigure[]{
  \includegraphics[width=0.40\textwidth]{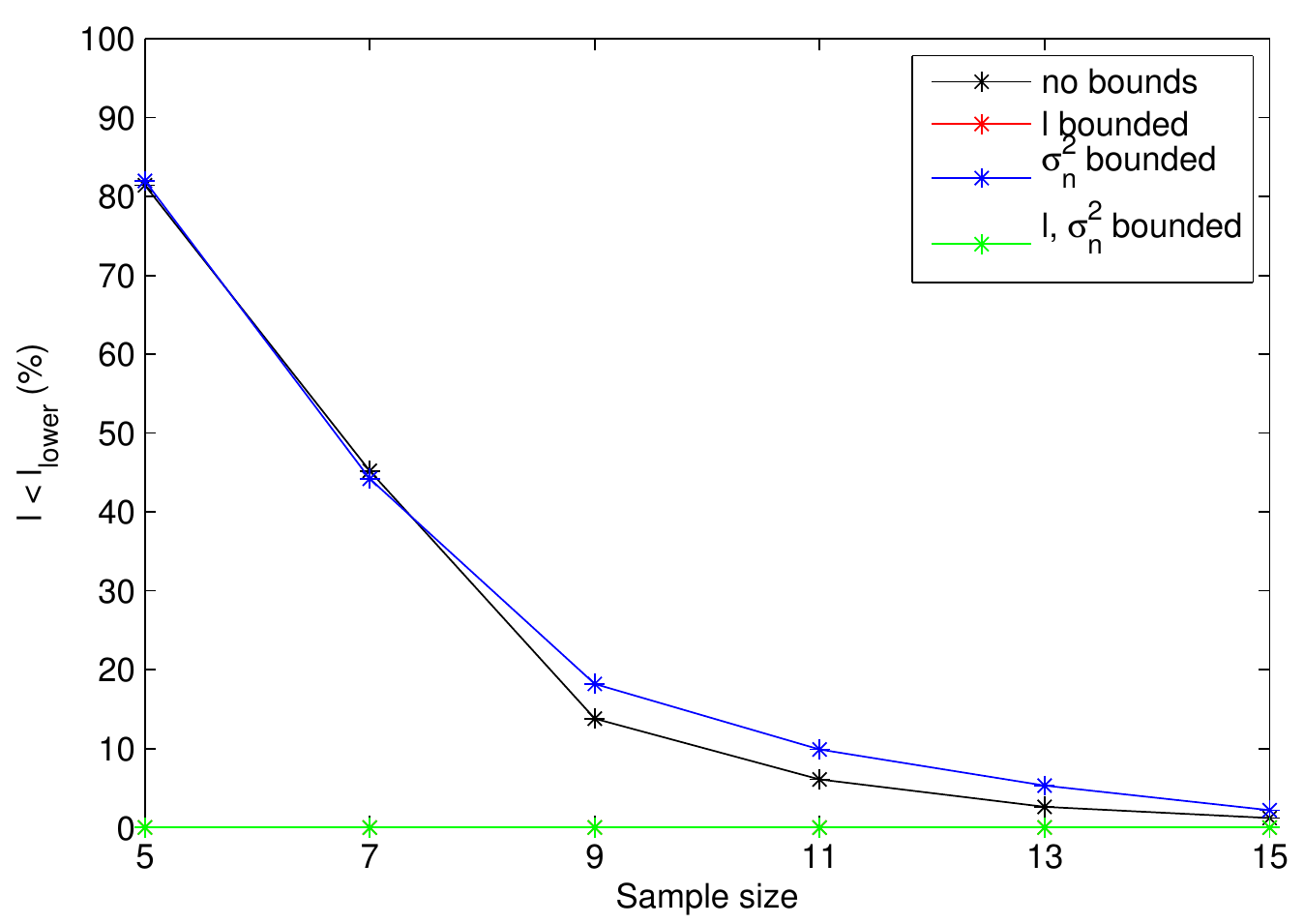}
  }
 \subfigure[]{
 \includegraphics[width=0.40\textwidth]{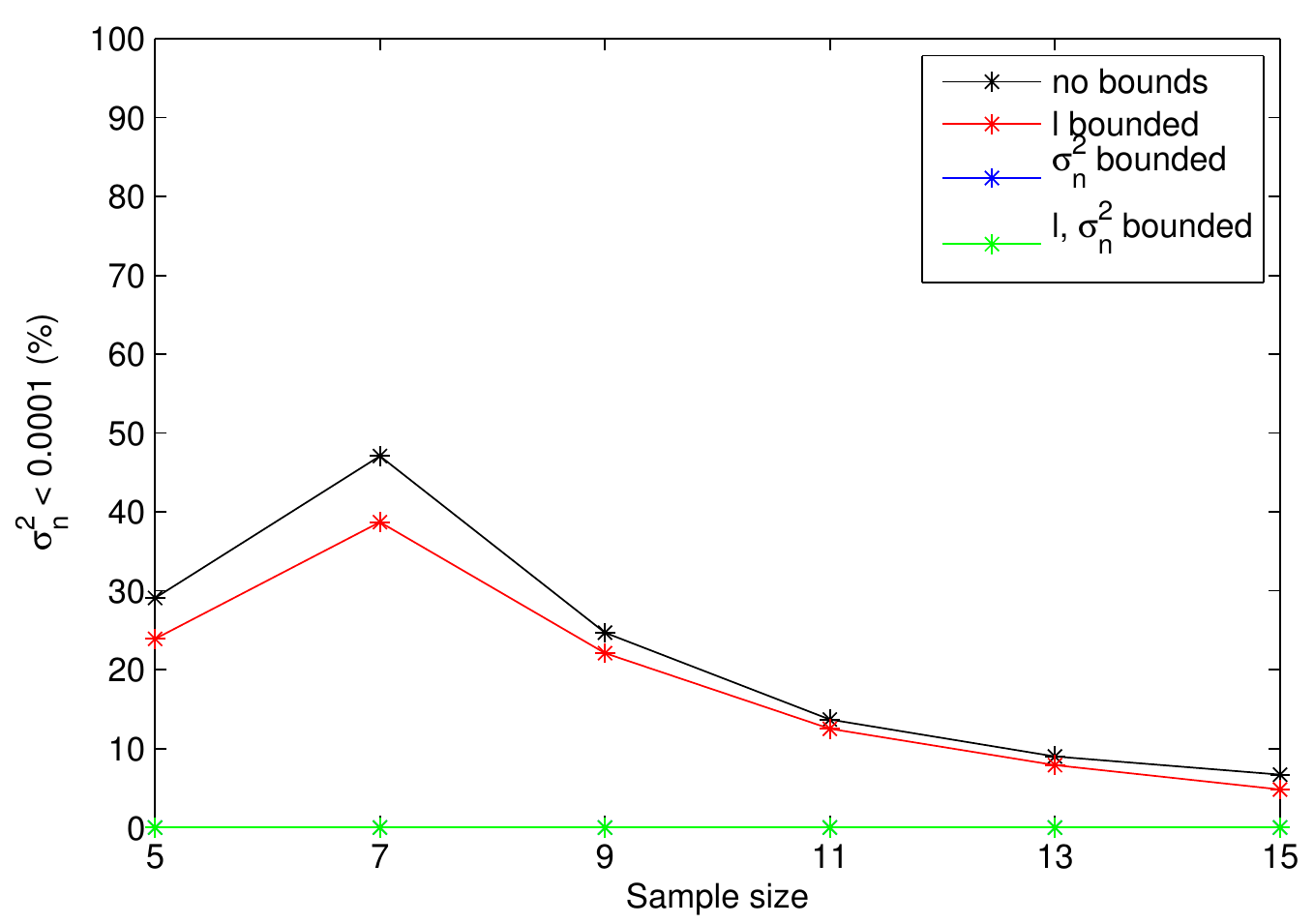}
 }
  \caption{Fraction of over-fitted models in different setups with
    synthetic data.  Setting the length-scale bound (a) automatically
    eliminates all problems with length scales and a noise bound (b) with
    the noise values, and the corresponding curves are overlapping at
    constant 0.}
  \label{fig:syn_data_results}
\end{figure*}

\begin{figure*}[tb]
  \centering
  \subfigcapskip-0.5ex
  \subfigcapmargin1ex
  \subfigure[]{
  \includegraphics[width=0.40\textwidth]{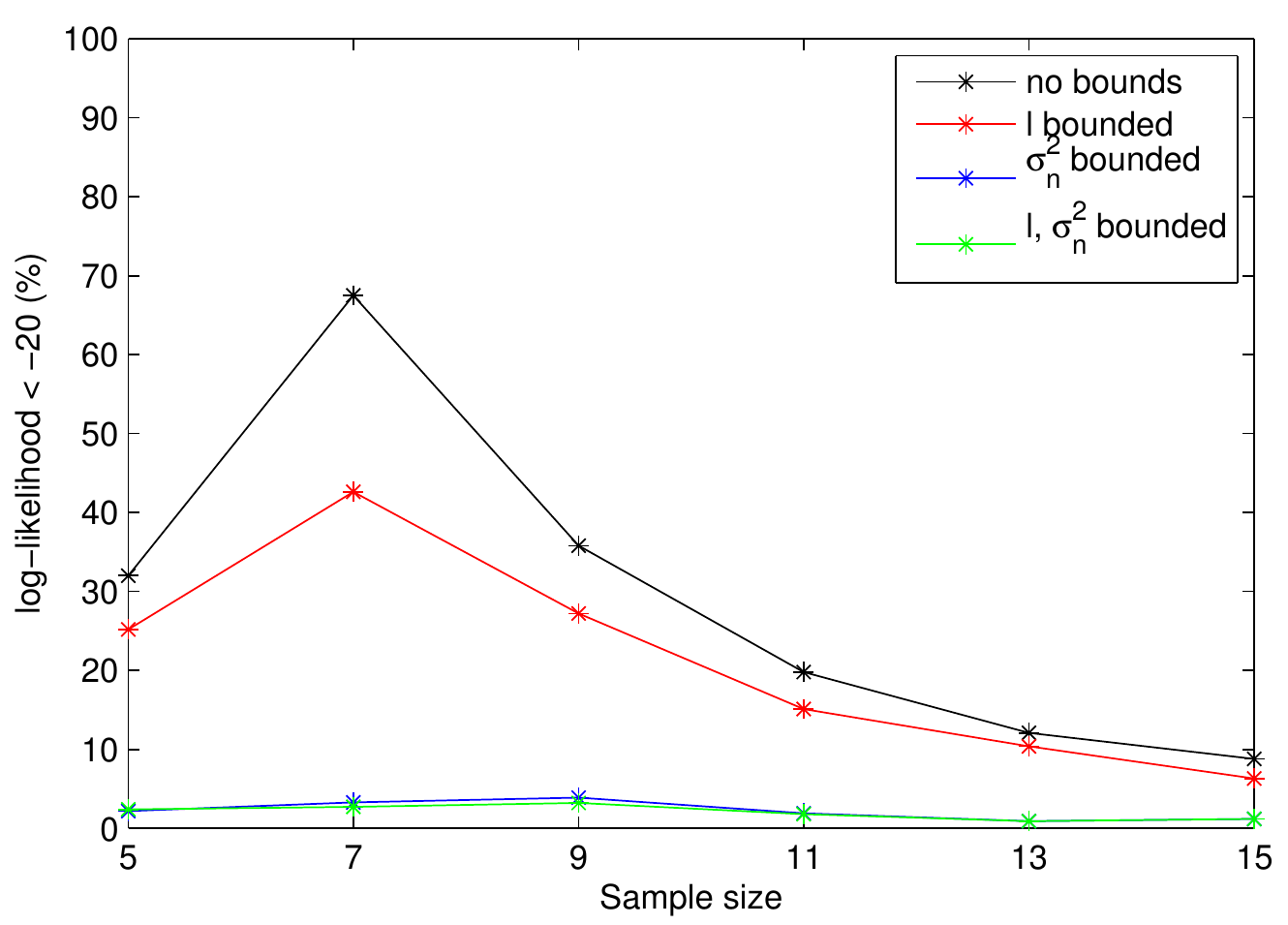}
  }
 \subfigure[]{
 \includegraphics[width=0.40\textwidth]{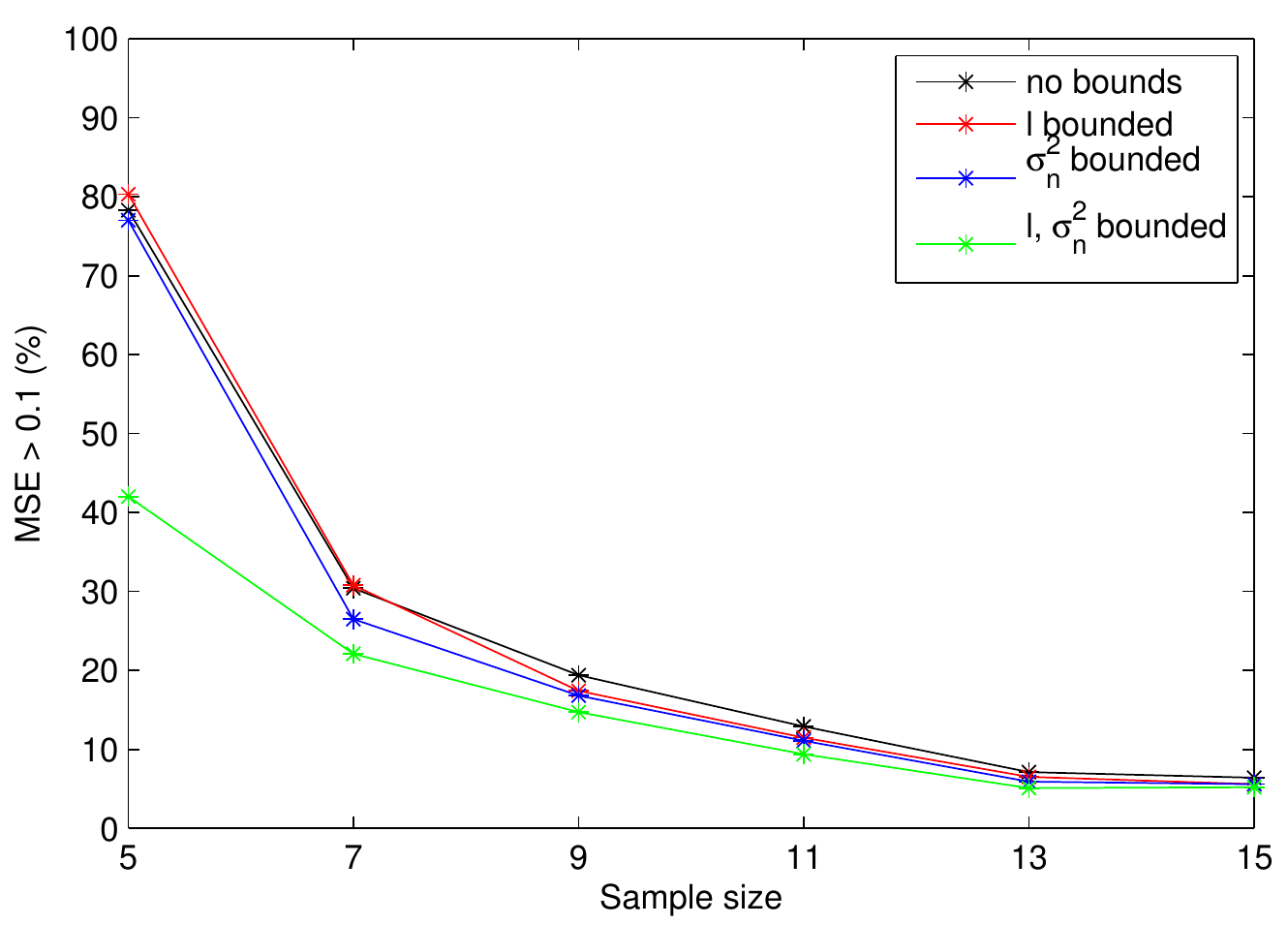}
 }
  \caption{Fraction of models with low predictive log-likelihood values (a) and with high MSE values (b) in different setups with synthetic data.}
  \label{fig:l_plot}
\end{figure*}

\begin{figure*}[tb]
  \centering
  \subfigcapskip-0.5ex
  \subfigcapmargin1ex
  \subfigure[No bounds]{
    \includegraphics[width=0.41\textwidth,trim=10mm 71mm 10mm 71mm,clip]{no_bound.pdf}
  }%
  \subfigure[Bounded $\ell$]{
    \includegraphics[width=0.41\textwidth,trim=10mm 71mm 10mm 71mm,clip]{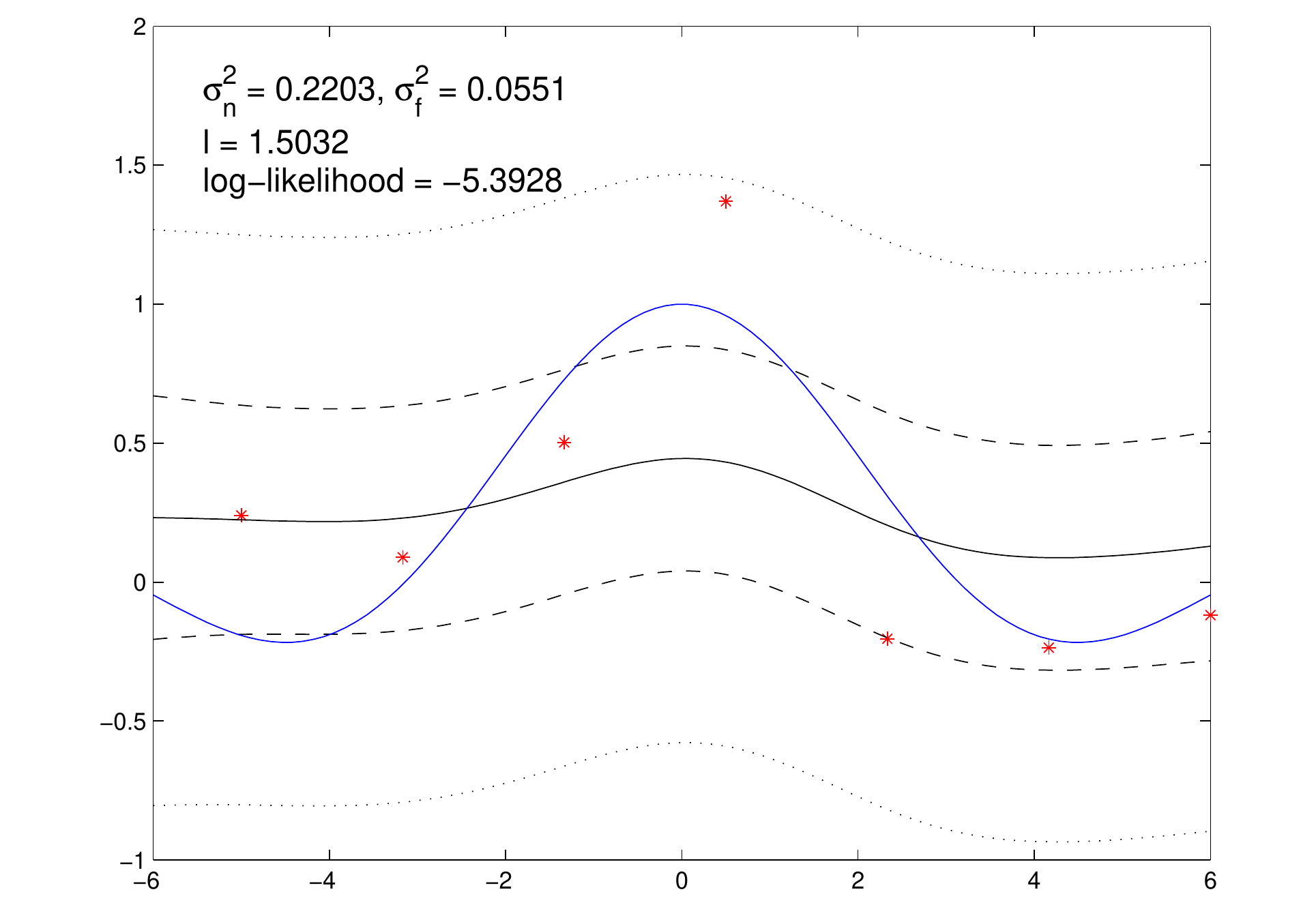}
  }
  \subfigure[Bounded $\sigma_n^2$] {
    \includegraphics[width=0.41\textwidth,trim=10mm 71mm 10mm 71mm,clip]{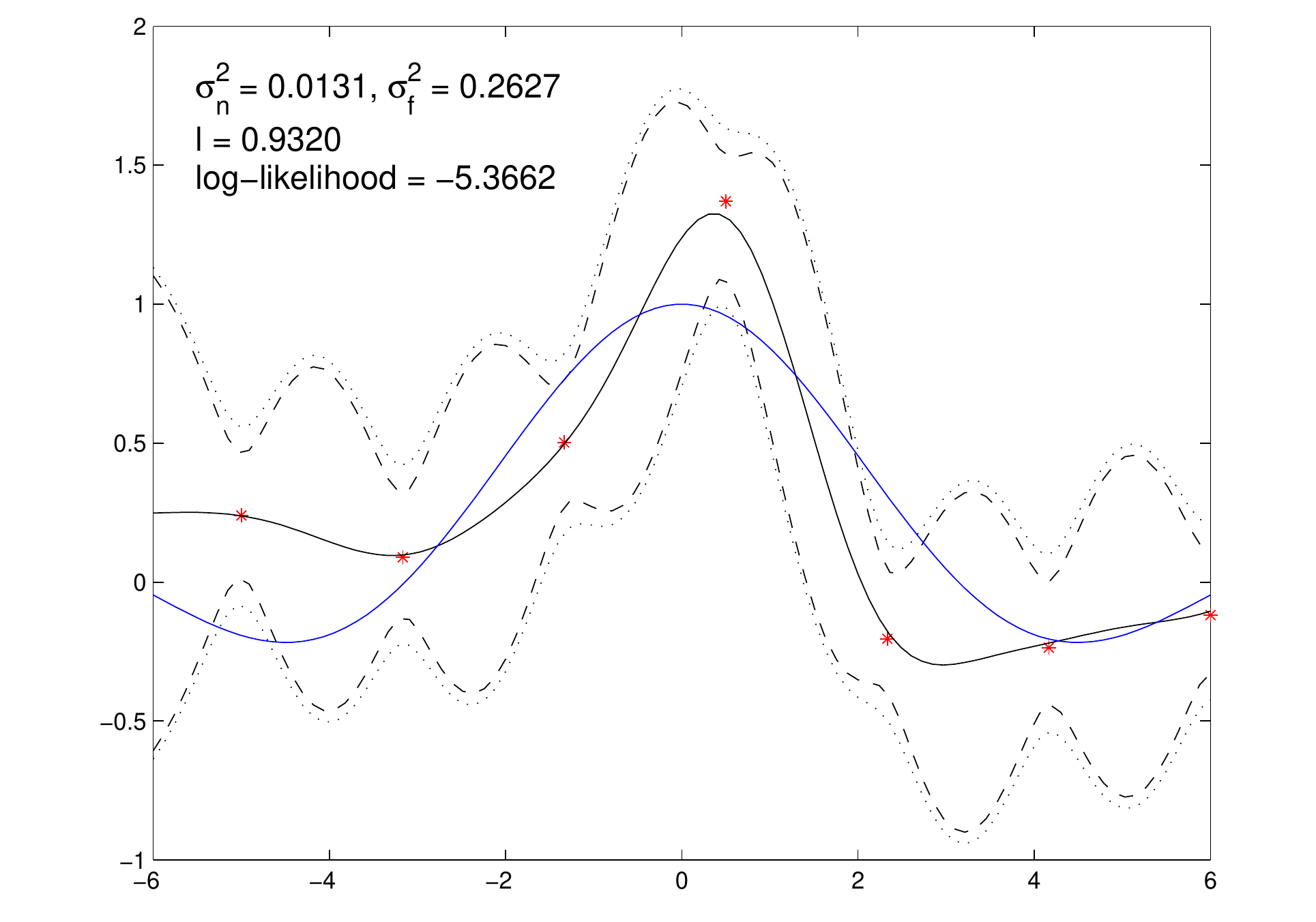}
  }
  \subfigure[Bounded $\ell$ and bounded $\sigma_n^2$] {
    \includegraphics[width=0.41\textwidth,trim=10mm 71mm 10mm 71mm,clip]{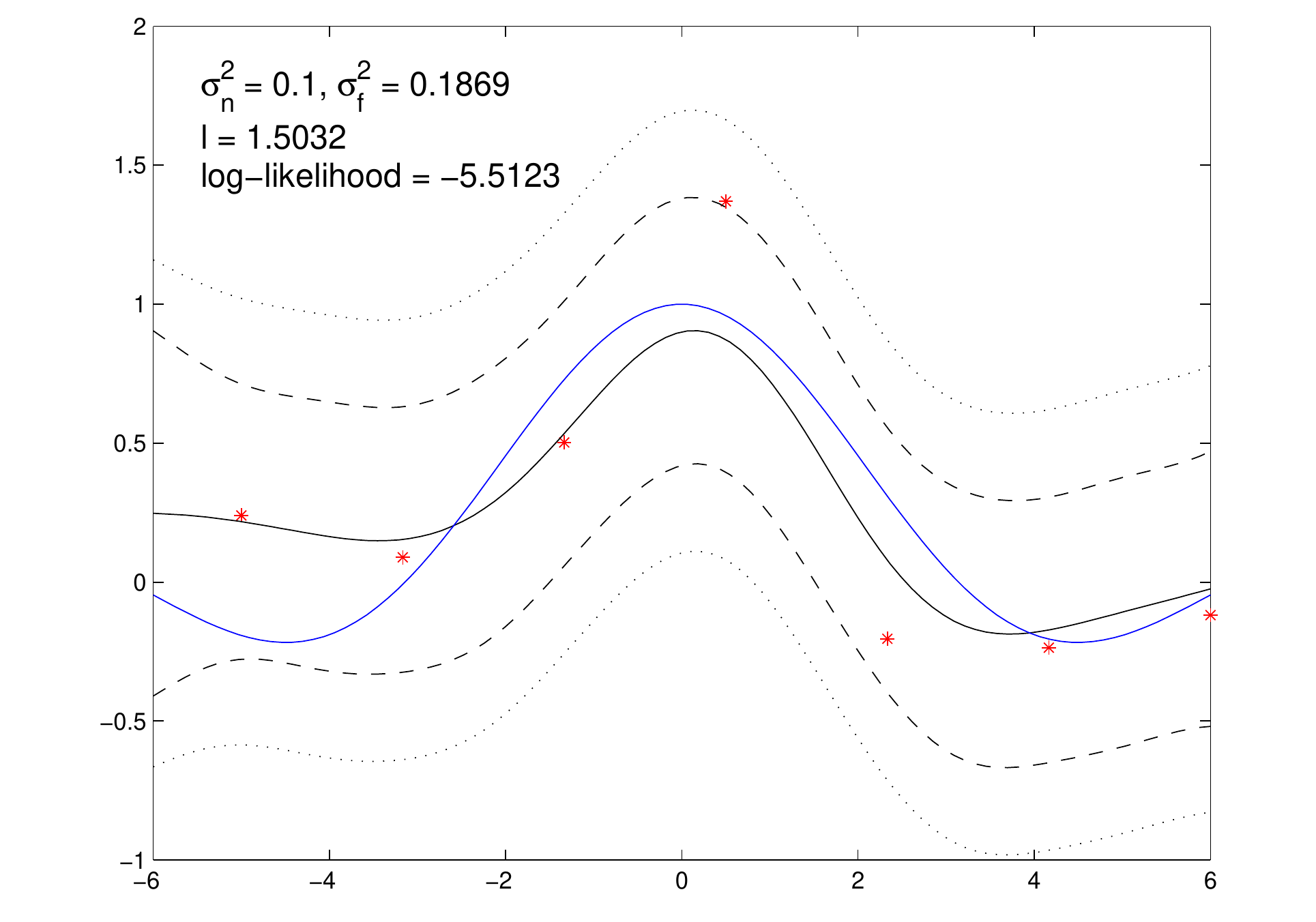}
  }
  \caption{Examples of GP model fits for a synthetic data instance.
    Using no bounds in (a) leads to severe over-fitting which
    is compensated by under-fitting by introducing the $\ell$ bound ($[\ellbound{}, \infty)=[1.5032, \infty)$, from $\Delta t \approx 1.8334$) in
    (b), while introducing only the $\sigma^2_n$ bound ($[0.01,0.1]$) does not help avoid over-fitting in (c).  Introducing the $\ell$ and $\sigma^2_n$ bounds together in (d) leads to a reasonably good fit to the original function.}
  \label{fig:synthetic_example}
\end{figure*}

\subsection{Synthetic data}

We generated synthetic data using a procedure similar to the one
described in Sec.~\ref{sec:illustrative-example}.  We created 1000
independent instances of the data set with different noise
realisations.  We repeated this sampling $n=5,7,\ldots,15$ equally
spaced points on the interval $[-5,6]$.

For each of these data sets, we fitted the model in four different
scenarios:
\begin{enumerate}
\item Unconstrained $\ell$ parameter estimation and noise $\sigma_n^2$ estimation;
\item Unconstrained $\sigma_n^2$ estimation and bounded $\ell$ in the range $[\ellbound{}, \infty)$;
\item Bounded $\sigma_n^2$ in the range $[0.01,0.1]$ and unconstrained $\ell$; and
\item Bounded $\sigma_n^2$ in the range $[0.01,0.1]$ and bounded $\ell$ in the range $[\ellbound{}, \infty)$.
\end{enumerate}
For each $n$ and each scenario, we recorded the number of fits with
$\ell < \ellbound{}$ and $\sigma_n^2 < 0.0001$, both of which are
indications of over-fitting.  The results are shown in
Fig.~\ref{fig:syn_data_results}.  For very small data sets,
practically all instances lead to a short length scale, and a
significant fraction has a very small estimated noise variance.  For
larger data sets the fractions drop, but remain well above zero.

Alternatively, to compare how the fitted GP models coincide with the true underlying function, we calculated the mean squared errors (MSE) by using 10 test points, equally spaced in the range [-6,5]. MSE values can be found simply by calculating the mean of the squared differences between the true underlying function values and the predicted function values at the test points. Additionally, we also computed the predictive log-likelihood values for the true function values at the test points. Small MSE values, and large log-likelihood values can be considered as the indicators of a good fit, whereas high MSE values and low log-likelihood values indicate the opposite.

The frequencies of the instances with very low predictive log-likelihood values (smaller than $-20$) and with very high MSE values (larger than $0.1$) can be seen in Fig.~\ref{fig:l_plot}(a) and Fig.~\ref{fig:l_plot}(b), respectively. It is clear that the frequencies are very high if no bounds are set to the parameters. Once the parameter bounds are introduced, the frequencies start to decrease, with the largest decrease occuring in the instances with small sample sizes. Furthermore, for each instance, we recorded in which setup the fitted GP model leads to the smallest MSE and the largest predictive log-likelihood values. In Table~\ref{max_loglik} and Table~\ref{min_mse}, the fractions of the largest predictive log-likelihood values and the smallest MSE values in four different setups are presented, supporting the fact that using the length scale and noise variance bounds together improves the model fit drastically especially in the instances with small sample sizes.

Model fits for an example realisation, with sample size 7, suffering from over-fitting with
unbounded estimation are shown in Fig.~\ref{fig:synthetic_example}.
The over-fitting in the unbounded estimate in
Fig.~\ref{fig:synthetic_example}(a) is clearly remedied by introducing
the length-scale bound in Fig.~\ref{fig:synthetic_example}(b), but
this makes the model under-fit.  Also constraining the noise variance to
sensible values in Fig.~\ref{fig:synthetic_example}(d) leads to a
reasonably good fit considering the amount of data.

\begin{table*}[htb]
  \centering
  \caption{Fractions ($\%$) of the largest log-likelihoods in different setups with synthetic data ($n$ denotes the sample size)}
  \vskip 0.15in
  \begin{center}
  \begin{small}
  \begin{sc}
  \begin{tabular}{|l|c|c|c|c|c|c|}
    \hline
    & $n=5$ & $n=7$ & $n=9$ & $n=11$ & $n=13$ & $n=15$ \\
    \hline
    \small
    No bounds
    & 2.2 & 3.0 & 9.6 & 10.6 & 11.9 & 11.6 \\
    \hline
    \small
    $\ell$ bounded
    & 1.0 & 8.5 & 11.6 & 14.9 & 13.7 & 13.0 \\
    \hline
    \small
    $\sigma^2_n$ bounded
    & 28.3 & 34.2 & 44.1 & 37.6 & 39.7 & 34.6 \\
    \hline
    \small
    $\ell$ and $\sigma^2_n$ bounded
    & 68.5 & 54.3 & 34.7 & 36.9 & 34.7 & 40.8 \\
    \hline
 \end{tabular}
 \end{sc}
\end{small}
\end{center}
\vskip -0.1in  
  \label{max_loglik}
\end{table*}

\begin{table*}[htb]
  \centering
  \caption{Fractions ($\%$) of the smallest MSEs in different settings with synthetic data ($n$ denotes the sample size)}
  \vskip 0.15in
  \begin{center}
  \begin{small}
  \begin{sc}
  \begin{tabular}{|l|c|c|c|c|c|c|}
    \hline
    & $n=5$ & $n=7$ & $n=9$ & $n=11$ & $n=13$ & $n=15$ \\
    \hline
    \small
    No bounds
    & 8.1 & 9.4 & 13.7 & 19.1 & 22.3 & 23.6 \\
    \hline
    \small
    $\ell$ bounded
    & 5.7 & 14.3 & 20.0 & 22.6 & 25.0 & 26.0 \\
    \hline
    \small
    $\sigma^2_n$ bounded
    & 27.0 & 36.2 & 38.8 & 28.9 & 26.5 & 21.5 \\
    \hline
    \small
    $\ell$ and $\sigma^2_n$ bounded
    & 59.2 & 40.1 & 27.5 & 29.4 & 26.2 & 28.9 \\
    \hline
 \end{tabular}
  \end{sc}
\end{small}
\end{center}
\vskip -0.1in  
  \label{min_mse}
\end{table*}

\begin{table*}[htb]
  \centering
  \caption{Proportion of over-fitted single-target cascaded
    differential equation models according to different criteria
    in different experimental setups. '.' indicates a situation that
    is not possible because of the bounds.}
    \vskip 0.15in
  \begin{center}
  \begin{small}
  \begin{sc}
  \begin{tabular}{|l|c|c|c|c|c|c|c|c|}
    \hline
    & \multicolumn{2}{|c|}{BIN}
    & \multicolumn{2}{|c|}{MEF2}
    & \multicolumn{2}{|c|}{TIN}
    & \multicolumn{2}{|c|}{TWI} \\
    \hline
    & $\ell < \ellbound{}$ & $\sigma_n^2 < 0.01$
    & $\ell < \ellbound{}$ & $\sigma_n^2 < 0.01$
    & $\ell < \ellbound{}$ & $\sigma_n^2 < 0.01$
    & $\ell < \ellbound{}$ & $\sigma_n^2 < 0.01$ \\
    \hline
    No bounds
    & 0.0\% & 0.0\%
    & 0.0\% & 0.3\%
    & 13.9\% & 2.9\%
    & 0.0\% & 1.6\% \\
    \hline
    $\ell$ bounded
    & . & 0.0\%
    & . & 0.3\%
    & . & 4.1\%
    & . & 1.6\% \\
    \hline
    $\sigma_n^2$ fixed
    & 1.8\% & .
    & 1.9\% & .
    & 5.3\% & .
    & 7.1\% & . \\
    \hline
    $\ell$ bounded, $\sigma_n^2$ fixed
    & . & .
    & . & .
    & . & .
    & . & . \\
    \hline
  \end{tabular}
  \end{sc}
\end{small}
\end{center}
\vskip -0.1in  
  \label{tab:single_target_res}
\end{table*}

\subsection{Gene expression data}
\label{sec:gene-expression-data}

We apply the model to fruit fly \emph{Drosophila melanogaster}
developmental gene expression time series from~\citep{Tomancak2002}
using the differential-equation-based gene regulation model
from~\citep{Lawrence2007,Gao2008,Honkela2010}.

The experiments were run using a modified version of the \emph{tigre}
Bioconductor package~\citep{Honkela2011}.  For each model, we tested 4
different scenarios:
\begin{enumerate}
\item Unconstrained $\ell$ parameter estimation and noise $\sigma_n^2$ estimation;
\item Unconstrained $\sigma_n^2$ estimation and bounded $\ell$;
\item Fixed $\sigma_n^2$ from pre-processing and unconstrained $\ell$; and
\item Fixed $\sigma_n^2$ and bounded $\ell$.
\end{enumerate}
For the last two, the noise variances were obtained from data
pre-processing as in~\citep{Honkela2010}.  The models were run for 6795
genes passing the significant activity filter described
in~\citep{Honkela2010}.

\paragraph{Single-target cascaded differential equation models}

As the first example, we tested single-target cascaded differential
equation models linking observed regulator TF mRNA to candidate target
mRNA.  We ran the model for 4 TFs: BIN, MEF2, TIN and TWI.  Fractions
of genes with $\ell < \ellbound{}$ or $\sigma_n^2 < 0.01$ for each TF
for each setting are listed in Table~\ref{tab:single_target_res}.

\begin{figure*}[tb]
  \centering
  \subfigcapskip-0.5ex
  \subfigcapmargin1ex
  \subfigure[No bounds]{
    \includegraphics[width=0.2\textwidth,trim=120mm 0mm 0mm 0mm,clip,page=1]{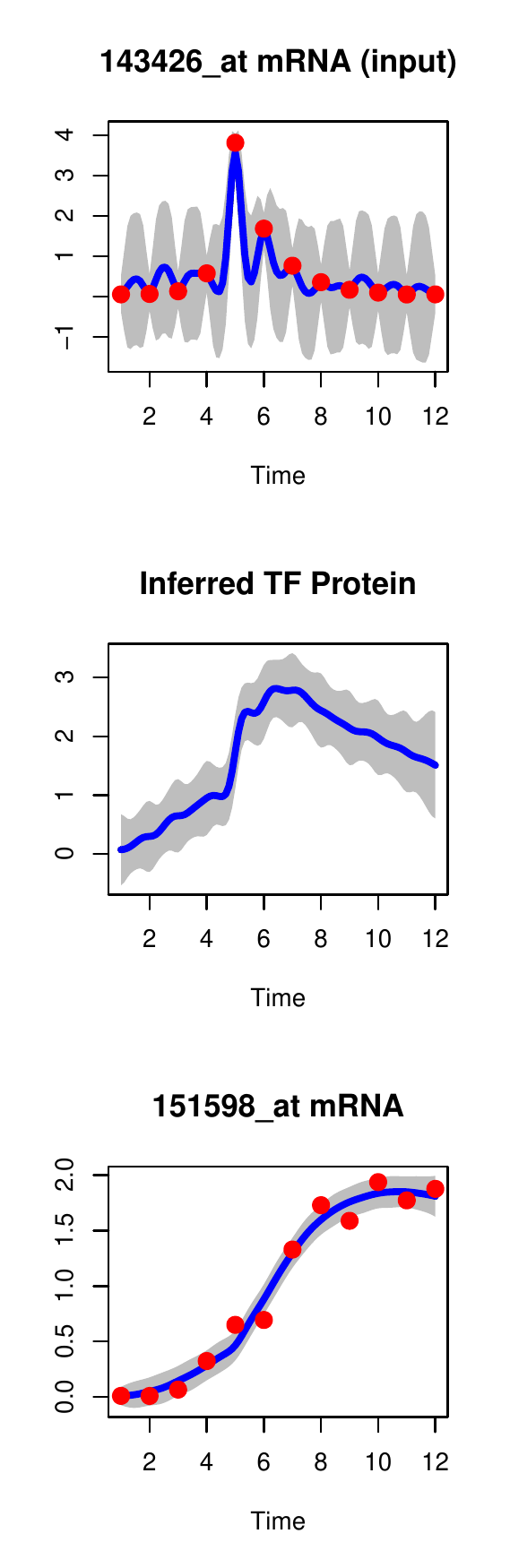}
  }%
  \subfigure[Bounded $\ell$]{
    \includegraphics[width=0.2\textwidth,trim=120mm 0mm 0mm 0mm,clip,page=2]{tin_single_target_model.pdf}
  }
  \subfigure[Bounded $\ell$ and fixed $\sigma_n^2$] {
    \includegraphics[width=0.2\textwidth,trim=120mm 0mm 0mm 0mm,clip,page=3]{tin_single_target_model.pdf}
  }
  \caption{Example of a single-target cascaded ODE gene regulation model.
    Using no bounds in (a) leads to severe over-fitting which is
    mostly corrected by $\ell$ bound in (b).  More accurate noise
    model in (c) further improves the accuracy slightly.}
  \label{fig:single_target_example}
\end{figure*}

The results show a moderate number of genes that exhibit symptoms of
over-fitting.  The numbers vary significantly for different TFs, as the
strength of the driving signal is different.  The expression input for
TIN is especially weak and sharply peaked, which leads to relatively
larger number of over-fitted models without the precautions proposed in
this paper.  An example of such a model is illustrated in
Fig.~\ref{fig:single_target_example}.

%%%%

%%%%

\paragraph{Multiple-target models}

As the second example we consider a slightly more difficult task of
fitting a single-layer differential equation model with no TF mRNA
input~\citep{Gao2008}.  We fit the model to each gene in turn with two
fixed known targets, MEF2 and TIN.  These are both known targets of
TWI~\citep{Zinzen2009}, so these models could plausibly be used to
discover further targets of TWI.  Each model was fitted independently
without using any information from the previous models.

Fractions of genes with different symptoms of over-fitting are listed
in Table~\ref{tab:multi_target_res}.  The numbers are in most cases
slightly higher than in Table~\ref{tab:single_target_res},
demonstrating that this is a more challenging task.

\begin{table}[h]
  \centering
  \caption{Proportion of over-fitted multiple-target non-cascaded
    differential equation models according to different criteria
    in different experimental setups.}
    \vskip 0.15in
  \begin{center}
  \begin{small}
  \begin{sc}
  \begin{tabular}{|l|c|c|}
    \hline
    & $\ell < \ellbound{}$ & $\sigma_n^2 < 0.01$ \\
    \hline
    No bounds
    & 7.3\% & 8.3\% \\
    \hline
    $\ell$ bounded
    & . & 6.7\% \\
    \hline
    $\sigma_n^2$ fixed
    & 6.7\% & . \\
    \hline
    $\ell$ bounded, $\sigma_n^2$ fixed
    & . & . \\
    \hline
  \end{tabular}
  \end{sc}
\end{small}
\end{center}
\vskip -0.1in  
  \label{tab:multi_target_res}
\end{table}

\section{Discussion}
\label{Discussion}

We have presented methods for improving large-scale learning of GP
models for very many instances of small data sets.  We presented a
novel rigorous derivation for a bound for sensible length scales for
squared exponential and Mat\'{e}rn covariance functions.  The bound is
based on constraining the energy spectrum of the GP covariance to
frequencies that can plausibly be reconstructed from the data.  This
can be intuitively justified by the fact that the data was collected
at such sampling rate in the first place, which encodes a prior
assumption about the time scale of interest.  This bound
clearly helps avoid many cases of obvious over-fitting, as illustrated
with both synthetic and real data experiments.

The usual underlying reason for the small length-scale fits is often
that a smooth model may not be a very good fit to the specific
instance of data.  The GP model may attempt to compensate for this by
using more functional degrees of freedom than seems plausible and
essentially modelling each single observation independently.  For
highly constrained models such as the linear ODE models in
Sec.~\ref{sec:gene-expression-data} the results are usually not very
severe, but for more flexible models such as ones incorporating
gene-dependent delays the over-fitting can be even more severe
problem.

One may wonder whether the observed small length-scales are caused by
using point estimates for the parameters.  This does not
appear to be the case, as illustrated by the likelihood surface in
Fig.~\ref{fig:likelihood_surface}(b) which shows a very smooth maximum in
small length-scale regime.  This is further supported by the fact that
some form of length-scale bounds was needed by~\citet{Titsias2012} for
a method completely based on MCMC.

We believe the presented length-scale bound is an important ingredient
for large-scale learning of multiple independent GP models for short time series
that are very common in biology. Carefully justified priors will be
especially important in the future when moving beyond the simple
linear ODE models of~\citet{Honkela2010} to more realistic and flexible
models.  Using such models without suitable constraints or suitably
constrained priors can easily lead to unexpected over-fitting failure
modes.

\small

\bibliography{smallgp}
\bibliographystyle{abbrvnat}

\end{document}